\documentclass[10pt,twocolumn,twoside]{IEEEtran} 

\usepackage{algorithm,algorithmic}
\usepackage{amsmath,graphicx}
\usepackage{epstopdf}
\usepackage{amssymb}
\usepackage{multirow}
\usepackage{epstopdf}
\usepackage{graphicx}
\usepackage{xcolor}
\usepackage{booktabs}

\def\Expect{\mathbb{E}}
\def\R{\mathbb{R}}
\def\N{\mathcal{N}}

\def\ltwo{\ell_2}
\def\lp{\ell_p}

\newtheorem{Theorem}{Theorem}

\newtheorem{Definition}{Definition}

\begin{document}
%
\title{Sample Distortion for Compressed Imaging}
%
%

\author{Chunli Guo\authorrefmark{1},~\IEEEmembership{Student Member,~IEEE,}
        and Mike E. Davies,~\IEEEmembership{Senior Member,~IEEE}\\
\thanks{The authors are with the Institute for Digital Communication and with the Joint Research Institute for Signal and Image
Processing, Edinburgh University, King’s Buildings, Mayfield Road, Edinburgh EH9 3JL, UK (e-mail: c.guo@ed.ac.uk, mike.davies@ed.ac.uk).

CG receives her PhD Scholarship from the Maxwell Advanced Technology Fund at the University of Edinburgh. MED acknowledges support of his position from the Scottish Funding Council and their support of the Joint Research Institute with the Heriot-Watt University as a component part of the Edinburgh Research Partnership in Engineering and Mathematics. 

Some of the work presented here was previously been presented in \cite{SD} and at the \textit{IEEE International Conference on Acoustics, Speech, and Signal Processing ICASSP 2013}, Vancouver, Canada.}}
\maketitle

\begin{abstract}
We propose the notion of a sample distortion (SD) function for independent and identically distributed (i.i.d) compressive distributions to fundamentally quantify the achievable reconstruction performance of compressed sensing for certain encoder-decoder pairs at a given sampling ratio. Two lower bounds on the achievable performance and the intrinsic convexity property is derived. A zeroing procedure is then introduced to improve non convex SD functions. The SD framework is then applied to analyse compressed imaging with a multi-resolution statistical image model using both the generalized Gaussian distribution and the two-state Gaussian mixture distribution. We subsequently focus on the Gaussian encoder-Bayesian optimal approximate message passing (AMP) decoder pair, whose theoretical SD function is provided by the rigorous analysis of the AMP algorithm. Given the image statistics, analytic bandwise sample allocation for bandwise independent model is derived as a reverse water-filling scheme. Som and Schniter's turbo message passing approach is further deployed to integrate the bandwise sampling with the exploitation of the hidden Markov tree structure of wavelet coefficients. Natural image simulations confirm that with oracle image statistics, the SD function associated with the optimized sample allocation can accurately predict the possible compressed sensing gains. Finally, a general sample allocation profile based on average image statistics not only illustrates preferable performance but also makes the scheme practical.


\end{abstract}


%
\IEEEpeerreviewmaketitle

\section{Introduction}
%
%
%
%

\PARstart{T}{raditionally} in compressed sensing (CS) a lot of work has been done in improving reconstruction algorithms assuming the optimality of the homogeneous random sensing matrix. There has recently been more attention on tailoring the sensing matrix in accordance with the signal of interest. We focus on designing a block diagonal measurement matrix for wavelet representation of natural images which falls under the general scope of bandwise sampling.

Donoho pioneered the use of bandwise sampling for compressed sensing in his original paper \cite{CS}. Tsaig further expanded the idea through the concept of two-gender CS, which randomly samples the fine-scale wavelet coefficients while fully samples in the coarse-scale domain \cite{Twogender}. In \cite{MBCS}, a specific sampling pattern is provided for the general multi-scale image model. With the key component of weighing the wavelet band importance, it achieves considerable improvement over the homogeneous matrix. However, the weight for each wavelet scale is assigned empirically. Despite all the attempts to improve the measurement matrix, the prior works are algorithmic and lack a solid theoretical grounding.

Analytically optimizing the bandwise sample allocation of the sensing matrix was originally considered in \cite{infosensing} and \cite{InforCS}. The authors seek to minimize the reconstruction uncertainty in terms of the entropy of the CS approximation. However, directly quantifying the entropy is very difficult, thus the authors resorted to an ad hoc solution, which only approximately optimizes the InfoMax criterion \cite{linsker1989application}.

In fact, the notion of optimized bandwise sampling dates back much further and was instrumental in Kashin's proof of the optimal rates of approximation (n-widths) for certain classes of smooth function \cite{Kashin1977}, which was a key inspiration for the theory of compressed sensing \cite{CS}. Specifically, bandwise sampling forms the basis of Maiorov's discretization theorem which relates function n-widths to the n-widths of a sequence of finite dimensional $\lp$ balls \cite{Maiorov1975}.

In other recent work, the block diagonal spatially-coupled sensing matrix was used to reach the fundamental undersampling limit of compressed sensing with almost perfect reconstruction \cite{MagicMatrix}, \cite{MagicMatrixExtend}. Unfortunately, to achieve the ground-breaking improvement, a good level of compressibility that we do not normally observe in natural images is required, which makes it impractical for compressed sensing of real images.
 
\subsection*{Main Contributions}
In this work, we seek to better understand the nature of good sample allocation strategies for multi-resolution images. To this end, we begin by setting up the sample distortion (SD) framework for a stochastic CS model. The SD function is proposed with the purpose of assessing the performance of different encoding and decoding methods quantitatively in terms of the expected mean squared error (MSE) distortion. Then an entropy based bound on the achievable MSE performance for any linear encoder-CS decoder pair is derived following the classic rate distortion theory. A tighter distribution specific model based bound is further derived by leveraging the entropy based bound of the Gaussian source. We then prove that the SD function is convex in nature. It comes with a key insight: any scheme whose SD function is concave over the sampling ratio interval $[0, \delta_c]$ can be improved for any $\delta$ in that interval, by sensing a portion of the source at the rate $\delta_c$ and making no attempt to sense the reminder. The zeroing procedure which can convexify the SD function comes naturally as a result. 

As a broad definition, the SD function is applicable to any encoder-decoder pair, i.e. the Gaussian homogeneous encoder with the linear $\ell_2$ decoder or the $\ell_1$ minimum CS decoder. In this work, we mainly investigate the SD function for the Bayesian optimal approximate message passing (BAMP) decoder \cite{AMP}, \cite{BayAMP}. In the context of the replica method, the BAMP decoder can be tuned for optimal performance and admits a rigorous analysis in the large system-limit with a large set of sub-Gaussian encoders, which naturally provides the theoretical basis for its SD function \cite{AMP_SE}, \cite{dynamicAMP}, \cite{ISIT2012Bayati}. Two compressible distributions: the generalized Gaussian distribution and the two-state Gaussian mixture distribution are selected as the representative examples, because they are commonly used models in the compressed imaging literature \cite{GMCrouse}, \cite{GMDuarte}, \cite{GGDMoulin}, \cite{GGDBouman}.

The second part of the paper makes a contribution to the understanding of analytically optimizing the per-band sample allocation for a bandwise independent image model. For this we use an orthogonal wavelet model to make sure our analysis tractable. We have proved that the optimal sample arrangement with the minimum MSE is achieved by performing a reverse water-filling strategy, given the per-band statistics and by virtue of the convexified SD function. A similar idea was used in \cite{infosensing} to design the sensing matrix that is most informative about the source. A water-filling strategy is also used in \cite{Schiniter2011CAMSAP} in the context of adaptive sensing. The reconstruction quality can be quantitatively predicted and evaluated by the SD function for the multi-resolution image model. Given the oracle image statistics, our SD function based sample allocation is the best we can achieve in terms of minimizing the MSE. In practice, where the image information is not always available, the performance depends on the quality of the image statistic estimation. 


Finally wavelet dependencies are incorporated with the bandwise sampling by modelling the wavelet coefficients with the hidden Markov tree (HMT) structure \cite{GMCrouse}. Several works have exploited the local dependencies of the wavelet coefficients in the wavelet based compressed sensing literature, such as \cite{MCMC}, \cite{kim2012wavelet} and \cite{averbuch2012adaptive}. In this paper we leverage Som and Schniter's state-of-the-art turbo message passing approach to alternate between the CS decoding and the tree structure decoding \cite{TurboAMP}. Instead of using a uniform distribution of samples across wavelet bands, we choose the optimized block diagonal sensing matrix to sample independently in the CS decoding procedure. We see that even with the sub-optimal sample allocation from the bandwise independent model, the exploitation of the wavelet tree structure enables the accurate message from the coarse scale bands propagates to the fine scale bands and eventually benefit the reconstruction. Attempts are made to find better sample allocation for the tree structure image model. Empirical results are obtained for a specific image example. However, finding the truly best sample allocation for the turbo method is beyond the scope of this paper.

The remainder of the paper is organized as follows. We set up the sample distortion framework in Section \ref{sec:2},  In Section \ref{sec:3} optimizing sample allocation for multi-scale wavelet image model and incorporating with tree structure is discussed. Simulation results are given in Section \ref{sec:4} and Section \ref{sec:5} concludes. 

\section{Sample Distortion Framework} 
\label{sec:2}

\subsection{SD Function Definition}
Suppose the signal of interest $\mathbf{x}\in\R^n$ is a random vector (source) with i.i.d components drawn according to the prior distribution $p(x)$. Denote $\boldsymbol{\Phi}\in\R^{m\times n}$, $m<n$ as the sensing matrix or the encoder, and $\mathbf{y}=\boldsymbol{\Phi}\mathbf{x}$ as the underdetermined linear combination of the source. Let $\delta=m/n$ be the sampling ratio. The goal of statistical compressed sensing is to reconstruct $\mathbf{x}$ using some Lipschitz regular mapping $\boldsymbol{\Delta} : \R^m\to\R^n$ based on the knowledge of $\mathbf{y}$, $\boldsymbol{\Phi}$ and $p(x)$. In our work, we are interested in the reconstruction quality for certain encoder-decoder pairs $(\boldsymbol{\Phi},\boldsymbol{\Delta})$ at an sampling ratio $\delta$, which is evaluated by the expected error distortion between the original signal $\mathbf{x}$ and the estimation $\boldsymbol{\Delta}(\boldsymbol{\Phi}\mathbf{x})$:

\begin{equation}
\label{eq:MSE}
D_{\lbrace\boldsymbol{\Phi},\boldsymbol{\Delta}\rbrace}(\delta)=\frac{1}{n}\Expect||\mathbf{x}-\boldsymbol{\Delta}(\boldsymbol{\Phi}_{\delta} \mathbf{x})||^2_2
\end{equation}

Along the lines of the classical rate-distortion function in the communication field \cite{shannon}, we define a sample distortion function for the compressed sensing setting.
\begin{Definition}
The Sample Distortion (SD) function is defined as the infimum of sampling ratios for which there is an encoder-decoder pair, $(\boldsymbol{\Phi},\boldsymbol{\Delta})$, that can achieve an expected distortion $D$.\end{Definition}

\begin{equation}
D(\delta)=\inf_{\boldsymbol{\Phi}, \boldsymbol{\Delta}, n} D_{\lbrace \boldsymbol{\Phi},\boldsymbol{\Delta}\rbrace}(\delta)
\end{equation}

Through a slight abuse of terminology, we will also use the term SD function to refer to the minimum distortion level a specific encoder-decoder pair can achieve at a fixed sampling ratio for a given compressive source. In this paper we will concentrate on the Gaussian encoder-BAMP decoder pair. 

\begin{figure}[t!]
 \centering
 \includegraphics[scale=0.5]{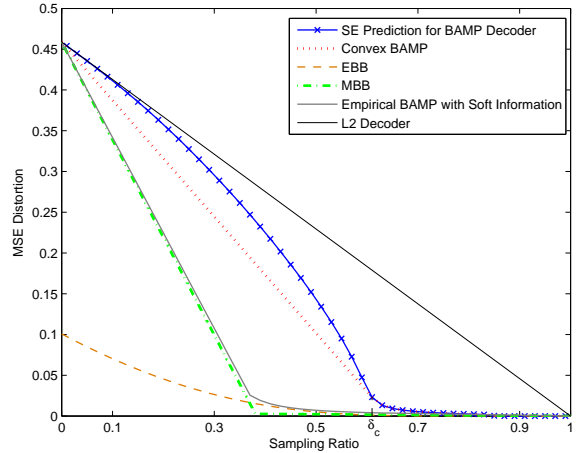}
 \caption{SD functions for GMD data $p(x)=0.38\;\N(0,1.198)+0.62\;\N(0,0.004)$ and lower bounds. The critical sampling ratio to convexify this SD function is $\delta_c=0.61$.}
 \label{fig:sdb3GM}
\end{figure}

\subsection{SD Function for BAMP}

Recent work by Donoho, Maleki and Montannari has shown that the AMP algorithm can achieve the same sparsity undersampling trade-off as the corresponding $\ell_1$ convex optimization procedure, but at less computational cost \cite{AMP}. When the signal prior, $p(x)$, is known, the general AMP algorithm can be tuned optimally by replacing the soft thresholding step with an optimal scalar MMSE estimator to improve the recovery algorithm \cite{BayAMP}, \cite{Bamp}. Moreover, the asymptotic MSE behaviour of BAMP can be precisely characterized by the state evolution (SE) formalism \cite{dynamicAMP}, known as the cavity method in the context of statistical physics \cite{MagicMatrixExtend} as $n\rightarrow\infty$, which allows us to make theoretical prediction about the SD performance of the Gaussian encoder-BAMP decoder. On the large-system limit assumption, the distortion iteration can be derived from the SE function \cite{Bamp}, \cite{MagicMatrixExtend} \footnote{While the large-system limit assumption does not hold, there is no analogous results like \eqref{eq:SEGM}. The finite-\textit{n} case has been studied in a recent work by Rangan et al. \cite{RanganFiniteNAMP}.}
\begin{equation}
\label{eq:SEGM}
D_{k+1}=\Expect\{[F(\tilde{\mathbf{x}}+\sqrt{\frac{D_{k}}{\delta}}\mathbf{z} ; \frac{D_k}{\delta})-\tilde{\mathbf{x}}]^2\}
\end{equation}
where $\tilde{\mathbf{x}}$ follows the choice of the compressive distribution, $\mathbf{z}\sim\N(0,1)$ is independent of $\tilde{\mathbf{x}}$, and $D_0=\Expect(\tilde{\mathbf{x}}^2)$. The function $F(;)$ is the (non-linear) scalar MMSE optimal estimator for $\tilde{\mathbf{x}}$ given $\tilde{\mathbf{x}}+\mathbf{z}$. The expectation in \eqref{eq:SEGM} is taken with respect to $\tilde{\mathbf{x}}$ and $\mathbf{z}$ and is in general calculated numerically. The SD function for BAMP decoder $D_{_\textrm{BAMP}}(\delta)$ is then given by the fixed point \footnote{For the distributions considered in this paper there is only one fixed point, i.e. BAMP exhibits no phase transitions} of \eqref{eq:SEGM}.

\begin{figure}[t!]
 \centering
 \includegraphics[scale=0.5]{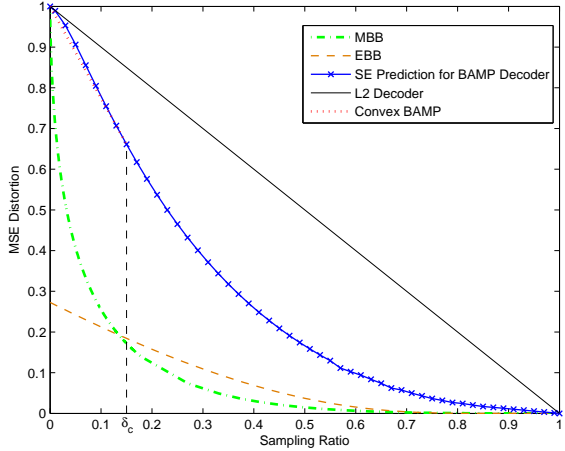}
 \caption{SD functions for GGD data $\alpha=0.4$, $\sigma=1$ and lower bounds. The critical sampling ratio to convexify this SD function is $\delta_c=0.15$.}
 \label{fig:sdggd}
\end{figure}

We will consider two specific non-Gaussian distributions, the two-state Gaussian Mixture distribution (GMD) and the generalized Gaussian distribution (GGD), to model the compressive random vector. 

As the combination of two Gaussian distributions with large variance $\sigma_{_L}^2$ and small variance $\sigma_{_S}^2$, the GMD model is quite effective at capturing the heavy tailed nature of an approximate sparse signal by adjusting the activity rate $\lambda$. A random vector with i.i.d GMD components can be seen as generated from either the small variance Gaussian distribution or from the large one, depending on the hidden states $s=\lbrace0,1\rbrace$.
\begin{equation}
\begin{split} 
p_{_\textrm{GMD}}(x)=& p(x, s=1)+ p(x,s = 0)\\ 
=& p(s=1)\N(x;0,\sigma_{_L}^2)\\
~&+p(s=0)\N(x;0,\sigma_{_S}^2)\\
      =&\lambda \N(x;0,\sigma_{_L}^2)+(1-\lambda)\N(x;0,\sigma_{_S}^2)
\label{eq:GMD}
\end{split}
\end{equation} 

Another popular probabilistic model for compressive data is the generalized Gaussian distribution (GGD). The pdf for the GGD can be written as
\begin{equation}
p_{_\textrm{GGD}}(x) = \frac{\alpha}{2 \sqrt{\beta} \sigma \Gamma(\frac{1}{\alpha})}
\exp \left( - \left| \frac{x}{\sqrt{\beta} \sigma}\right|^\alpha \right)
\end{equation}
where $\beta = \Gamma(1/\alpha)/\Gamma(3/\alpha)$, $\sigma$ is the standard deviation and $\alpha$ is the shape parameter. As $\alpha$ goes to zero the distribution has increasingly heavy tails. For images we are typically interested in the GGD with $\alpha\sim [0.3,1]$ since these distributions provide a good approximation for the distribution of the wavelet coefficients in a given band for natural images. 

Examples of the theoretical prediction for the SD function of GMD and GGD data using BAMP decoder can be found in Fig. \ref{fig:sdb3GM} and Fig. \ref{fig:sdggd} respectively. The function $F(;)$ has a close-form expression for the GMD \cite{MagicMatrixExtend}, \cite{TurboAMP} and can be solved numerically for the GGD.

\subsection{SD Lower Bound}
\label{sec:LB}
To understand the fundamental theoretical limits of CS for compressible distributions, we now derive two lower bounds for the SD function. We first prove the \textit{entropy based bound} (EBB) which is a sampling analogy to the classical Shannon Rate Distortion Lower Bound (this result first appeared in the conference paper \cite{SD}). 
\begin{Theorem}
Let $\mathbf{x}\in\R^n$ be a realization of the random vector $\mathbf{x}=x_1,\cdots,x_n$, $i.i.d.\sim p(x)$, Var$(x_i)=1$ and $h(x_i)<\infty$. Let $\mathbf{y}=\boldsymbol{\Phi}\mathbf{x}$, $\mathbf{y}\in\R^{\delta n}$, $\delta=m/n<1$. Then for any Lipschitz reconstruction decoder $\boldsymbol{\Delta} : \R^{m}\to\R^n$, we have:
\begin{equation}
D_{\boldsymbol{\Delta}}(\delta)\geq(1-\delta)2^{2(h(x)-h_g)/(1-\delta)}
\end{equation}
where $h_g=\frac{1}{2}\log_2 2\pi e$ is the entropy of a unit variance Gaussian random variable.
\label{T:EBB}
\end{Theorem}

The proof is given in appendix \ref{app:EBB}. 

The EBB can be easily rescaled to bound the SD performance for distributions with non-unit variance. When the source $\mathbf{x}$ is Gaussian the second term in the lower bound becomes 1. The EBB for a Gaussian distribution reduces to the well known form: $D_{_\textrm{EBB}}=1-\delta$, which can be shown to be tight. Furthermore when we use the linear estimator (optimal for Gaussian source), $\hat{\mathbf{x}}=\boldsymbol{\Phi}^{\dagger}\mathbf{y}$, it is straight forward to show that the SD function, $D_{\ell_2}(\delta)$, has the same form independent of the pdf of the source $\mathbf{x}$ and is achievable with any full rank linear encoder. 

While the EBB in Theorem \ref{T:EBB} provides a bound on the achievable performance of CS specifically for i.i.d sources, it is not clear how close we can expect to get to it. The EBB for both GMD and GGD data are plotted in Fig. \ref{fig:sdb3GM} and Fig. \ref{fig:sdggd}. We can see that at low sampling ratios, it is not tight since we expect the SD function, i.e. the MSE, to approach the signal energy as $\delta\to 0$.

We then define the \textit{model based bound} (MBB) to compensate for the disadvantage of the EBB. Inspired by the fact that the EBB is tight and achievable for Gaussian source, we resort to the hierarchical Bayesian model to approximate the target compressible distributions. By introducing the variance as a latent variable, the hierarchical representation of a compressive distribution $p(x)$ can be understood as the weighted sum of (possibly infinite) Gaussian distributions.
\begin{equation}
\begin{split}
\label{eq.MBB}
p(x)=&\int_0^{\infty} \! p(x|\tau)p(\tau) \, \mathrm{d}\tau \\
    =&\int_0^{\infty} \! \N(x;0,\tau)p(\tau) \, \mathrm{d}\tau
\end{split}
\end{equation} 

where $p(\tau)$ is the weight for the Gaussian component $\N(x;0,\tau)$. The MBB is then derived in the following manner: assume the source $\mathbf{x}$ is partitioned into different groups according to the variance. For both encoder and decoder, we agree to transmit and reconstruct the source group by group in the descendant order of the variance. For each Gaussian group, the SD function is tightly bounded by its EBB. Then the lower bound for the whole procedure can be seen as the weighted combination of the EBB of Gaussian components. Thus the MBB has the form:
\begin{equation}
D_{_\textrm{MBB}}(\delta)=\int_0^{c} \! \tau p(\tau) \, \mathrm{d}\tau
\end{equation} 
with $\delta=\int_c^{\infty} \! p(\tau) \, \mathrm{d}\tau$. 

The GMD model is intrinsically a discretized hierarchical Bayesian model with only two Gaussian components. Thus its MBB can be seen as the discretized version of the general form in :
\begin{equation}
D_{_\textrm{MBB}}(\delta)=\begin{cases}
(1-\lambda)\sigma_S^2+(\lambda-\delta)\sigma_L^2 & 0\le\delta\le\lambda\\
(1-\delta)\sigma_S^2 & \lambda<\delta\le 1
\end{cases}
\end{equation}
For the GGD model, the detailed procedure for inferring its hierarchical Bayesian prior $p(\tau)$ is relegated to appendix \ref{app:MBB}. As we can see in both Fig. \ref{fig:sdb3GM} and Fig. \ref{fig:sdggd}, the MBB is much tighter than the EBB for small sampling ratios, although neither the MBB nor the EBB dominates for the whole range of the sampling ratios. The supremum of the two therefore yields a better lower bound for the SD function.  

\subsection{Convexity Property}
In this subsection, we first prove that the SD function is necessarily convex. A direct application of this property is then illustrated to effectively improve the reconstruction quality of the Gaussian encoder-BAMP decoder in the low sample ratio regime.
\begin{Theorem}
The SD function $D(\delta)$ is convex.
\label{T:cvx}
\end{Theorem}

\begin{IEEEproof} Consider two achievable SD points ($\delta_1$, $D(\delta_1)$) and ($\delta_2$, $D(\delta_2)$). To prove the SD function is convex, we only need to show the convex combination of the two points is also achievable. Let $\delta_{t}=t\delta_1+(1-t)\delta_2$, $0\leq t\leq 1$. To sample the source $\mathbf{x}\in \R^n$ at the sampling ratio $\delta_t$, we could split $\mathbf{x}$ into two parts $\mathbf{x}=[\mathbf{x}_1,\mathbf{x}_2]^T$, where $\mathbf{x}_1\in\R^{tn}$, $\mathbf{x}_2\in \R^{(1-t)n}$, and apply encoders with sampling ratio $\delta_1$, $\delta_2$ to $\mathbf{x}_1$, $\mathbf{x}_2$, respectively. Then the reconstruction of $\mathbf{x}_1$ and $\mathbf{x}_2$ has achievable MSE: $tnD(\delta_1)$ and $(1-t)nD(\delta_2)$. So the MSE of the reconstruction of $X$ is:
\begin{equation}
nD(\delta_t)\leq tnD(\delta_1)+(1-t)nD(\delta_2)
\end{equation}
Therefore 
\begin{equation}
D(t\delta_1+(1-t)\delta_2)\leq tD(\delta_1)+(1-t)D(\delta_2)
\end{equation}
\end{IEEEproof}

\begin{figure}
 \centering
 \includegraphics[scale=0.6]{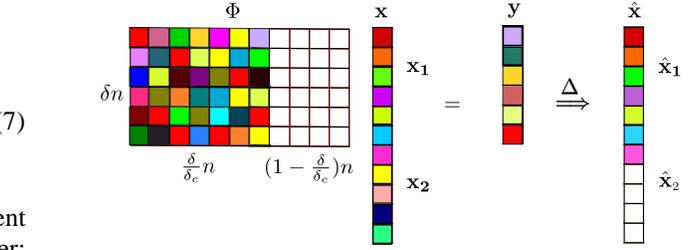}
 \caption{Hybrid zeroing Gaussian matrix as the convex combination of a trivial decoder $\hat{\mathbf{x}}=0$ and a BAMP decoder $\boldsymbol{\Delta}$. Elements equal to 0 are represented with white blocks.}
 \label{fig:cvx}
\end{figure}

The convexity property is applicable to the SD function for any specific encoder-decoder pair in the large-system limit. A direct consequence of Theorem \ref{T:cvx} is that for a given encoder-decoder pair with a concave SD function between $\delta_1$ and $\delta_2$ ($\delta_1<\delta_2$), there exists a hybrid system with better SD performance: it can be easily achieved by applying the two encoder-decoders to different portions of the source to get the convex combination of $D(\delta_1)$ and $D(\delta_2)$. A special case is when $\delta_1=0$ with the corresponding trivial decoder ($\hat{\mathbf{x}}=0$) and $\delta_2 = \delta_c$ with $\delta_c$ being the crucial sampling ratio. In this case, instead of sampling the source $\mathbf{x}$ with a full Gaussian matrix,
$\boldsymbol{\Phi}\in\R^{\delta n\times n}$, we
split $\mathbf{x}$ as before with $\mathbf{x}_1 \in \R^{tn}$ and
$\mathbf{x}_2 \in \R^{(1-t)n}$, $t = \delta/\delta_c$. We then sample
$\mathbf{x}_1$ with the Gaussian matrix, $\tilde{\boldsymbol{\Phi}}\in\R^{\delta n\times tn}$ and reconstruct, while the remaining $\mathbf{x}_2$ we
reconstruct as zero. Since this is equivalent to setting part of the encoder to
zero, $\boldsymbol{\Phi} = [\tilde{\boldsymbol{\Phi}},\boldsymbol{0}]$, we call this the \textit{zeroing procedure}, as illustrated in Fig. \ref{fig:cvx}.  

Close observation of the SD functions for the Gaussian encoder-BAMP decoder system in Fig. \ref{fig:sdb3GM} and Fig. \ref{fig:sdggd} reveals that the curves are convex for large sampling ratios but concave for small sampling ratios. By applying the hybrid zeroing Gaussian matrix, we convexify the SD function for $\delta$ below the crucial sampling ratio $\delta_c$. To best improve the SD performance, $\delta_c$ is chosen as the largest sampling ratio that below which the SD function is concave.

%


The Gaussian sensing matrix has been widely assumed within the CS community to be optimal in terms of CS performance. Indeed this has been proved to be the case for the distributions that exhibit exact sparsity \cite{WuVerdu}. However, under the assumption that the BAMP achieves the Bayes optimal reconstruction - this would follow, for example, if the replica method could be proved to be rigorous \cite{MagicMatrixExtend} - then the \textit{zeroing procedure} resulting from Theorem \ref{T:cvx} indicates this assumption to be false. It also shows that the optimal encoder is related to both the signal property and the corresponding decoding method. 

\section{Sample Distortion Function for Statistical Image Model}
\label{sec:3}
In this section we build upon the the aforementioned SD framework and study the SD behaviour of the compressive imaging. We investigate the optimal bandwise sampling strategy with a fixed sample budget, in a similar manner to \cite{infosensing}, but in terms of minimizing the expected MSE. We begin by introducing the bandwise independent multi-resolution statistical model for natural images. 

Natural images are transform compressible: they have more concise representation in the wavelet domain. The wavelet decomposition of an image $f(\mathbf{X})$ has the form \cite{waveletTour}:
\begin{equation}
f=\sum_k \mu_{i,k}\phi_{i,k}+\sum_{j\ge i,k}\omega_{j,k}\psi_{j,k}
\end{equation}
where $\phi_{i,k}$ are the scaling functions, $\psi_{j,k}$ are the prototype bandpass functions such that together they form an orthonormal basis. The variables $\mu_{i,k}$ are in turn the scaling coefficients at scale $i$ and $\omega_{j,k}$ are the wavelet coefficients at scale $j$. We can group the coefficients into a single vector according to the scale or band ${\boldsymbol{\theta}}=[\boldsymbol{\underline{\mu}_i}, \boldsymbol{\underline{\omega}_i}, \boldsymbol{\underline{\omega}_{i+1}},\cdots]^T$ and assign each a band index. For simplicity $\boldsymbol{\underline{\mu}_i}$ is band 0, the coarsest wavelet coefficients group, $\boldsymbol{\underline{\omega}_i}$, is denoted as band 1, and the rest can be labelled in the same manner. Here we follow \cite{ChoiBarniuk}, \cite{MomentMatching} and consider a simple statistical model defined directly on the wavelet coefficients. The band 0 is always treated as Gaussian since these coefficients typically exhibit no sparsity. This can be seen as a worse case assumption in terms of its SD function. For the other bands, we model the wavelet coefficients within each band as mutually independent and impose a compressive distribution for each wavelet band. To be specific, $\omega_{j,k}$ at scale $j$ can be modelled as  
\begin{equation}
\omega_{j,k}\sim \textrm{GGD}(0,\sigma_j^2,\alpha_j)
\end{equation}
or
\begin{equation}
\omega_{j,k}\sim \textrm{GMD}(\lambda_{j}, \sigma_{L,j}^2, \sigma_{S,j}^2),
\end{equation}
where typically for natural images the distributions exhibit a self-similar structure with an exponential decay across scale, i.e. $\sigma_j^2=2^{-j\beta}\sigma_0^2$ for the GGD and 
$\sigma_{a,j}^2 =2^{-j\beta}\sigma_{a,0}^2$, $a =S,L$  for the two-state GMD for some $\beta>0$. For the bandwise independent image model, we assume an uniform activity rate $\lambda_j$ for each wavelet band in spite of the coefficient index. In particular, we define $\lambda_j:=\textrm{Pr}\lbrace s_{j,k}=1\rbrace$.  

\subsection{Bandwise Sampling}
\label{sec:bandwiseSA}
To keep things tractable we restrict ourselves to the class of linear encoders, $\mathbf{y}=\boldsymbol{\Phi}\boldsymbol{\theta}$, that are block diagonal and sample the different wavelet bands separately with the following form:
\begin{equation}
  \boldsymbol{\Phi} =
  \begin{pmatrix}
  \boldsymbol{\Phi}_0 & & & \\
   & \boldsymbol{\Phi}_{1} & &\\
   & & \ddots &\\
   & & & \boldsymbol{\Phi}_{_L}
  \end{pmatrix}
    \end{equation}
where $\boldsymbol{\Phi}_{i} \in \R^{m_{i}\times n_{i}},m_i\le n_i$ puts $m_i$ measurements to sample the $i$th band. The equality holds when the $i$th band is fully sampled with $\boldsymbol{\Phi}_{i}$ being an identity matrix. Otherwise $\boldsymbol{\Phi}_{i}$ is a possibly zero padded (for convexity) Gaussian random matrix. And $\mathbf{\underline{y}}_i=\boldsymbol{\Phi}_i\boldsymbol{\underline{\omega}}_i$ is the CS observation for each block. To derive the SD function for the multi-resolution images, we first consider the $L$ wavelet bands as independent and parallel. The question then is how to allocate a fixed number of samples to the various bands, with the aim of minimizing the total reconstruction distortion. Let us assume for now that $m_i$, $n_i$ be continuous and $\delta_i=m_i/n_i\in[0,1]$. The problem is reduced to the following optimization
\begin{equation}
\begin{aligned}
\label{eq:IMSD}
& \min_{m_i} \sum_{i=1}^{L} \sigma_i^2 n_i D_i(m_i/n_i) \\
& \text{s.t. } \sum_{i=1}^{L} m_i=m \; \text{and} \; 0 \leq m_i \leq n_i, \; i = 1, \ldots, L. \\
\end{aligned} 
\end{equation}
where $D_i$ is the (convex) SD function for band $i$ normalized to have unit variance. Using Lagrange multipliers, we construct the objective function
\begin{equation}
\begin{split}
L=&-\sum_i \sigma_i^2 n_i D_i(m_i/n_i)\\
&-{\lambda}(\sum_i m_i-m)\\
&-\sum_i \mu_i(m_i-n_i)\\
&+\sum_i \nu_i m_i
\end{split}
\end{equation}

Differentiating with respect to $m_i$ and setting equal to 0 we have
\begin{equation}
\frac{\partial L}{\partial m_i}=-\sigma_i^2 n_i \frac{\partial D_i}{\partial\delta_i}\cdot\frac{\partial\delta_i}{\partial m_i}-{\lambda}-\mu_i+\nu_i=0
\end{equation} 
or
\begin{equation}
-\sigma_i^2\frac{\partial D_i}{\partial\delta_i}-{\lambda}-\mu_i+\nu_i=0
\end{equation}
Define the distortion reduction function as
\begin{equation}
\eta_i(\delta_i)=-\sigma_i^2\frac{\partial D_i}{\partial\delta_i},
\end{equation}
noting that this function is non-increasing in terms of $\delta_i$.
Now applying the Kuhn-Tucker (KT) conditions we arrive at:
\begin{equation}
  \eta_i(\delta_i)-{\lambda}-\mu_i+\nu_i  = 0,
\end{equation}
with
\begin{equation}
  \mu_i(n_i-m_i)=0,~\mu_i  \geq 0,
\end{equation}
and
\begin{equation}
  \nu_i m_i = 0,~\nu_i \geq 0.
\end{equation}


We therefore have three cases for the distortion reduction
function. First, if $0 < m_i < n_i$ then $\mu_i = \nu_i = 0$ and the
sampling ratio, $\delta_i$, is set so that $\eta_i(\delta_i) =
\lambda$. Next suppose that $m_i = n_i$ so that $\delta_i = 1$. In
this case, the KT conditions imply that 
\begin{equation}
\eta_i(\delta_i)\geq \lambda, ~\forall \delta_i
\end{equation}
In the final case we have $m_i = 0$ and $\delta_i = 0$. Here the KT
conditions imply:
\begin{equation}
\eta_i(\delta_i)\leq \lambda, ~\forall \delta_i
\end{equation}
 
This gives us an optimal sample allocation strategy which is similar
to the reverse water-filling idea in rate distortion theory
\cite{Cover2006}. We allocate samples to the band with
the greatest distortion reduction value until another band has a
greater one or that band has been fully
sampled. The procedure is stopped when the total distortion reaches
the desired level.  

\begin{figure}
 \centering
 \includegraphics[scale=0.5]{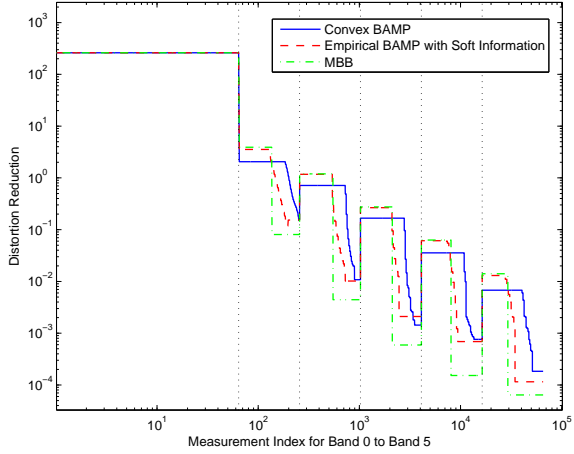}
 \caption{Distortion reduction function of six bands Daubechies 2 wavelet decomposition of cameraman image using GMD model (including the low-pass band). The statistics is reported in Table \ref{table:ImPara}.}
 \label{fig:DRF}
\end{figure}

To apply this idea to natural images we need to take account of the
fact that $m_i$, $n_i$ and $L$ are all discrete and finite. Thus we
define a discretized distortion reduction (DR) function for each
wavelet band. 
\begin{equation}
\eta_i(m_i)=\sigma_i^2[D_i(m_i/n_i)-D_i((m_i+1)/n_i)]
\end{equation} 
Suppose that $m_i$ samples have been allocated to the $i$th band. The
DR function gives the amount of distortion decreased by adding one
more sample to that band. Then the number of samples allocated to the
band $i$ is  
\begin{equation}
\label{eq:SA}
m_i=\begin{cases}
0 &\mbox{if } \max\eta_i(m_i)<\kappa\\
n_i & \mbox{if } \min\eta_i(m_i)>\kappa\\
\hat{m}_i \mbox{ s.t. }\eta_i(\hat{m}_i)=\kappa& \mbox{otherwise }
\end{cases}
\end{equation}
where $\kappa$ is chosen so that $\sum_i m_i=m$. With a convex
SD function, the optimal allocation is again achieved by performing a greedy
sample allocation strategy. The DR function for a six-band Daubechies 2 decomposition of the "cameraman" using the two-state GMD model is illustrated in Fig. \ref{fig:DRF}. One thing worth noting is that neither the convexity property nor the resulting greedy sample allocation method is restricted to the form of the decoder. For example the optimized bandwise sensing matrix can be designed in the same manner for the CS $\ell_1$ decoder and the $\ell_2$ decoder.




\subsection{Comparison to the Theory of Widths}
\label{subsec:nwidth}
In \cite{ChoiBarniuk}, parallels are drawn between the statistical wavelet model we have considered here and the family of Besov function spaces. In particular, the authors argue that under appropriate conditions realizations drawn from the GMD or GGD based wavelet model almost surely lie in an associated Besov space. It is therefore interesting to explore the similarities and differences between the achievable distortion rates derived here and those known in the deterministic setting for Besov spaces.

\subsubsection{n-widths of Besov spaces }

Consider the Lipschitz class of $r$-smooth functions on the interval $[0,1]$ and the unit ball, $B_p^r$, defined as:
\begin{equation}
B_p^r := \{f : \|f^{(r)}\|_p \leq 1\}
\end{equation}
where $f^{(r)}$ denotes the $r$th derivative of $f$ and the $L_p$ ball acts as the deterministic counterpart to the coefficient prior above. 

The $\ell_2$ error of the best n-dimensional linear approximation for these spaces is known to scale as $\sim n^{-r+1/p-1/2}$  for $1<= p <= 2$ \cite[Chapter 14, Theorem 1.1]{Lorentz1996}. In contrast, the $\ell_2$ error for the best CS reconstruction is characterized by the Gelfand width of $B_p^r$ which can be written as:
\begin{equation}
d^n(B_p^r) := \inf_{\boldsymbol{\Phi}} \sup_{h} \{\| h\|_2, h \in {\cal N}(\boldsymbol{\Phi}) \cap B_p^r\}.
\end{equation}
and measures the uncertainty in $B_p^r$ within the null space of $\boldsymbol{\Phi}$. Here, for $1 \leq p \leq 2$ the best CS approximation error decays at the faster rate of $\sim n^{-r}$, i.e. inversely proportional to the smoothness \cite[Chapter 14, Theorem 1.1]{Lorentz1996}. This result was derived in Kashin's seminal paper \cite{Kashin1977}, which is better known in the CS community for accurate bounds for the n-widths of $l_p$ balls in $\R^n$. 

\subsubsection{Similarities and differences}

Interestingly Kashin's result relied on a discretization theory of Maiorov \cite{Maiorov1975} that uses a similar bandwise sampling to our own. Specifically Maiorov uses a subband decomposition of spline spaces to bound the n-width of $B_p^r$ in terms of a weighted sum of finite dimensional n-widths for the individual subbands - effectively performing a bandwise sampling. Furthermore in both the deterministic and stochastic settings the allocation scheme is broadly the same: fully sample the first few low resolution subbands; then partially sample a number of intermediate subands; and finally set coefficients of all the higher resolution subbands to zero. However, in Kashin's theory, the number of partially sampled subbands grows as the distortion decreases and, indeed, it is this that accounts for the different rate of approximation compared with the best linear approximation. In contrast, in the sample allocation framework, the number of partially sampled bands, $P$, is bounded by the range of the distortion reduction function:
\begin{equation}
P < \beta \log_2 (\eta(0)/\eta(1)).
\end{equation}
For the two-state GMD model this bound is finite since from the MBB we can deduce that: 
\begin{equation}
\frac{\eta(0)}{\eta(1)} < \frac{\sigma_{L,0}^2}{\sigma_{S,0}^2}
\end{equation}
Note the same bound applies to the SD function for the MBB oracle decoder where the bandwise sampling is optimal. Hence, the fact that we do not get a growing number of partially sampled subbands implies that in the large system limit the CS approximation error will decay at the same rate as for the best linear approximation. We can therefore conclude that the gains in CS solutions over optimal linear approximation for such a model are fundamentally limited. We can see this, for example, in Fig. \ref{fig:DRF} where we would only ever partially sample at most 3 subbands for the convexified BAMP decoder.

\subsection{Incorporating Tree Structure}
\label{sec:TurboTree}
\begin{figure}
 \centering
 \includegraphics[scale=0.25]{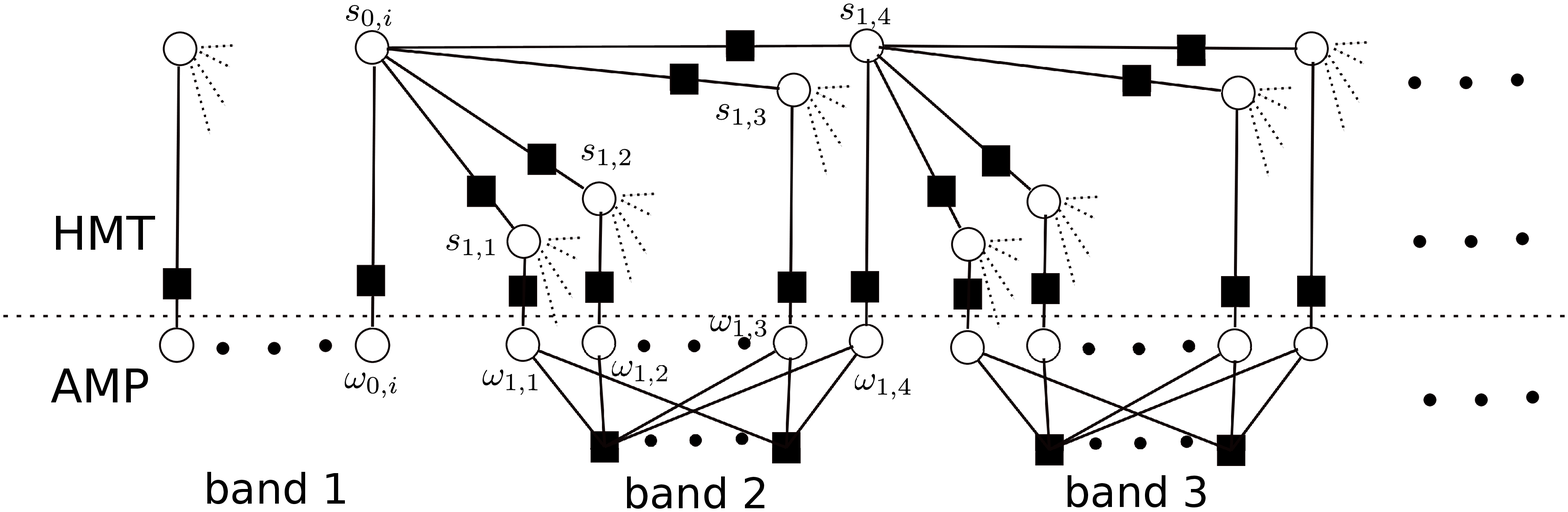}
 \caption{Factor graph for bandwise sampling with HMT decoding. The upper graph illustrates a quad-tree structure of the wavelet hidden states. The lower graph is the bandwise independent random mixing.}
 \label{fig:FG}
\end{figure}

Until now we have developed an analytic sample allocation method for a multi-resolution image model by assuming the independence of the wavelet band. In this subsection we look beyond the signal sparsity and incorporate the wavelet dependencies with the aim of getting closer to the upper model based bound. We model the wavelet coefficients with the GMD and impose the hidden Markov tree (HMT) structure to the hidden states as in \cite{GMCrouse}. To be specific, with the hidden states at the coarsest scale (band 1) being the "root", we connect each wavelet hidden state to the four "child" wavelet states one scale below it to form the image quad-tree (see HMT in Fig. \ref{fig:FG}). The \textit{persistence across scale} property \cite{waveletTour} states that the activity rate $\lambda_{j,k}$ for $\omega_{j,k}$ depends on the activity rate of its parent on scale $j-1$, $\lambda_{j-1,p_k}$, and the transition probabilities across scales. 
\begin{equation}
\begin{split} 
\lambda_{j,k}=& p(s_{j}=1|s_{j-1}=1)\lambda_{j-1,p_k}\\ 
~&+p(s_i=1|s_{j-1}=0)(1-\lambda_{j-1,p_k})\\
\label{eq:HMT}
\end{split}
\end{equation}

We first review the core principles of Som and Schniter's TurboAMP decoding method \cite{TurboAMP}. Let $\boldsymbol{\omega}=[\boldsymbol{\underline{\omega}_1}, \boldsymbol{\underline{\omega}_2}, \cdots, \boldsymbol{\underline{\omega}_L}]^T$ denote the collection of the wavelet coefficients of different bands and $\mathbf{s}=[\mathbf{\underline{s}_1}, \mathbf{\underline{s}_2}, \cdots, \mathbf{\underline{s}_L}]^T$ be the corresponding hidden states vector. Assume $\mathbf{y}=[\mathbf{\underline{y}_1}, \mathbf{\underline{y}_2}, \cdots, \mathbf{\underline{y}_L}]^T$ is the CS observation vector using the block diagonal sensing matrix. In the Bayesian compressed sensing setting, the reconstruction of $\boldsymbol{\omega}$ from $\mathbf{y}$ is interpreted as approximating the posterior mean of the density $p(\boldsymbol{\omega}|\mathbf{y})$:
\begin{equation}
\begin{split}
p(\boldsymbol{\omega}|\mathbf{y})&=Z^{-1}p(\boldsymbol{y}|\boldsymbol{\omega})\sum_{\mathbf{s}}p(\mathbf{s})p(\boldsymbol{\omega}|\mathbf{s})\\
&=Z^{-1}\sum_{\mathbf{s}}p(\mathbf{s})\prod_j[\prod_t p(y_{j,t}|\boldsymbol{\underline{\omega}}_j)][\prod_k p(\omega_{j,k}|s_{j,k})]
\end{split}
\end{equation}
where $Z=p(\mathbf{y})$. The factor graph plotted in Fig. \ref{fig:FG} visualizes this global function \cite{GraphicModelAMP}, \cite{FactorGraph}. Exact computation of $p(\boldsymbol{\omega}|\mathbf{y})$ is hard due to the dense and loopy structure of the factor graph. Instead we split the factor graph along the dashed line into two subgraphs as in \cite{TurboAMP} and calculate the marginal posterior $p(\omega_{j,k}|\mathbf{\underline{y}_j})$. The essence of turbo decoding is to exchange the local belief of the hidden states $s_{j,k}$ between AMP decoding and HMT decoding alternately, by treating the likelihood on $s_{j,k}$ from one decoding procedure as prior for the other decoding procedure. Here, unlike \cite{TurboAMP}, the AMP decoder is bandwise independent due to the block diagonal form of $\boldsymbol{\Phi}$. The interaction across different wavelet bands only comes from the HMT decoding.

The SD function for the bandwise independent image model is not optimal for the turbo decoding scenario since it does not take the HMT decoding into consideration. The role of the HMT decoding is to  better provide estimation of the activity rate $\lambda_{j,k}$ for the scalar MMSE estimator of each wavelet coefficient, instead of using an identical $\lambda_j$ over the coefficient index $k$, thus improving the reconstruction quality. To see the impact of the HMT decoding, we feed the BAMP decoder with the \textit{soft information}, $\hat{\lambda}_{j,k}$, defined as follows: 
\begin{equation}
\begin{split}
\hat{\lambda}_{j,k}=&\frac{p(\omega_{j,k}|s_{j,k}=1)}{p(\omega_{j,k}|s_{j,k}=1)+p(\omega_{j,k}|s_{j,k})=0}\\
=&\frac{\N(\omega_{j,k};0,\sigma^2_{j,L})}{\N(\omega_{j,k};0,\sigma^2_{j,L})+\N(\omega_{j,k};0,\sigma^2_{j,S})}
\label{eq:ExInfo}
\end{split}
\end{equation}
This provides a soft estimate of the state of the GMD and thereby gives a better prediction of individual coefficient variances. The empirical SD curve for BAMP decoder with soft information is generated from the Monte Carlo simulations with synthetic GMD data and illustrated in Fig. \ref{fig:sdb3GM}. To be specific, we use the $\hat{\lambda}_{j,k}$ in \eqref{eq:ExInfo} instead of $\lambda_j$ for the scalar MMSE estimator of each synthetic GMD component. Fig. \ref{fig:sdb3GM} demonstrates that providing the BAMP decoder with good estimation of activity rate information dramatically improves the reconstruction quality, with the SD function lying very close to the lower bound. 

Based on the per-band image statistics, the SD function for BAMP decoder with soft information can be obtained empirically for each wavelet band in the same fashion. Then the DR function with soft information for multi-resolution image model can be established following the aforementioned definition, as shown in Fig. \ref{fig:DRF}. To clarify the terminology, we denote the corresponding sample allocation profile as the HMT based sample allocation, or HSA. And we use the term SA to denote the sample allocation derived from the bandwise independent wavelet model. We should note here that neither SA nor HSA is optimal for turbo decoding. The problem with SA is that it tends to undersample the fine scale bands since they contain less energy than the coarse bands when treated independently. While HSA is served as the benchmark by assuming we have the accurate activity rate information for each wavelet coefficient. The optimal sample allocation for turbo decoding should combine the merits of both SA and HSA. 


\section{Natural Image Example}
\label{sec:4}
Reconstruction performance for natural images with the bandwise sampling matrix introduced in Section \ref{sec:3} is demonstrated and compared with several existing sensing matrices in this section. We start with the $256\times 256$ cameraman image as an introductory example. With the knowledge of the image statistics, we show that the bandwise independent image model based SD function can accurately predict the reconstruction quality for the proposed sample allocation scheme. It also confirms the theoretical optimality of our bandwise sensing matrix. We then extend the scheme to practical compressive imaging by designing the general sample allocation with the average image statistics estimated from the training set of the Berkeley dataset \cite{Berkeley}. Simulation with ten images from the test set further confirms that with good statistics estimation, the proposed SD sample allocation exhibits state-of-the-art performance. 

\subsection{Sample Allocation with Oracle Image Statistics}
The cameraman image is decomposed into six bands using the Daubechies 2 wavelet. GGD and GMD model parameters estimated directly from the wavelet coefficients are reported in Table \ref{table:ImPara} as the oracle image statistics, using moment matching \cite{MomentMatching} and EM algorithm \cite{EM} respectively. Given the parameter estimation, we are able to generate the image SD function and the subsequent bandwise sample allocation using the aforementioned method. 

\begin{table}[h!]
\caption{Statistics for Daubechies 2 wavelet coefficients of cameraman}\label{table:ImPara}
\vspace{-0.4cm}
\begin{center}
\resizebox{0.5\textwidth}{!}{
\begin{tabular}{c|c||cccccc}\hline
\multicolumn{2}{c}{subband}& $b_0$ & $b_1$ & $b_2$ & $b_3$ & $b_4$ & $b_5$\\ \hline \hline
\multirow{2}{*}{GGD}&$\alpha$ &2&0.7&0.4&0.3&0.3&0.4\\ \cline{2-8}
&$\sigma^2$ & 261.4383 & 2.0822 & 0.4559 & 0.0902 & 0.0167 & 0.0033 \\ \hline
\multirow{3}{*}{GMD}&$\lambda$ &1&0.4155&0.5309&0.4842&0.3664&0.2792\\ \cline{2-8}
&$\sigma_L^2$ &  261.4383 & 4.4215 & 0.8542 & 0.1856 & 0.0453 & 0.0115\\ \cline{2-8}
&$\sigma_S^2$ & & 0.3331 & 0.0038 & 0.0004 & 0.0002 & 0.0001\\ \hline
\end{tabular}}
\end{center}
\end{table}

To show the sample allocation method is not restricted to the form of the decoders, we consider three reconstruction options: the linear $\ell_2$ decoder, the CS $\ell_1$ decoder, and the BAMP decoder. The SPGL1 toolbox \footnote{http://www.cs.ubc.ca/labs/scl/spgl1/index.html} is used to implement the $\ell_1$ decoder. Its SD function can also be derived using the SE formalism \cite{dynamicAMP}. Both the $\ell_2$ and the $\ell_1$ decoder are considered for the GGD and the GMD model. Although in \cite{GMBAMPschinter} the authors show that the BAMP decoder is applicable to the GGD data by approximating it with the finite term of Gaussian mixture distribution, the approximation error may contribute to the final reconstruction distortion. Thus the BAMP decoder results are only reported for the GMD model here. The detailed algorithm can be found in \cite{TurboAMP}, \cite{MagicMatrixExtend}.

For quantitative comparison, the peak signal-to-noise ratio (PSNR) is used for both theoretical prediction and simulations. We examined the cameraman image at four different sampling ratios: $10\%$,$15.26\%$,$25\%$ and $30\%$ associated with $m=6554$, $10000$, $16384$, $19661$ noiseless measurements. Two different wavelet image models are considered. First, the wavelet bands are assumed as mutually independent. The proposed SA matrix is compared with five sensing matrices: the homogeneous Gaussian matrix (Uniform), the two-gender matrix (2 Gender) \cite{Twogender}, the informative sensing matrix (InforSA) \cite{InforCS} and the multi-scale sensing matrix (MBSA) in \cite{MBCS}. The 2 Gender matrix is implemented as fully sampling the scaling band and uniformly allocating the remaining samples to all the wavelet bands. As a statistic-dependent sample allocation scheme, InforSA is also generated based on Table \ref{table:ImPara}.

\begin{figure}
\centering
\includegraphics[width=0.45\textwidth]{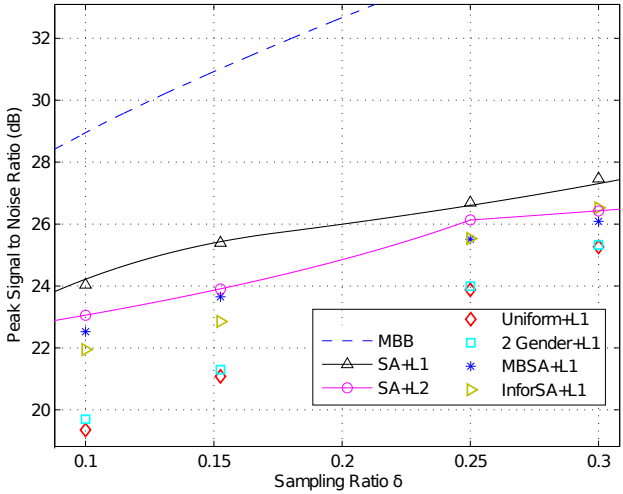}
\caption{PSNR comparison of different encoder-decoder pairs for cameraman Daubechies 2 wavelet with the GGD model. The lines are theoretical predictions with the SD function. While dots represent simulations with the cameraman image.}
\label{fig:SDRdb2GGD}
\end{figure}

\begin{figure}
\centering
\includegraphics[width=0.45\textwidth]{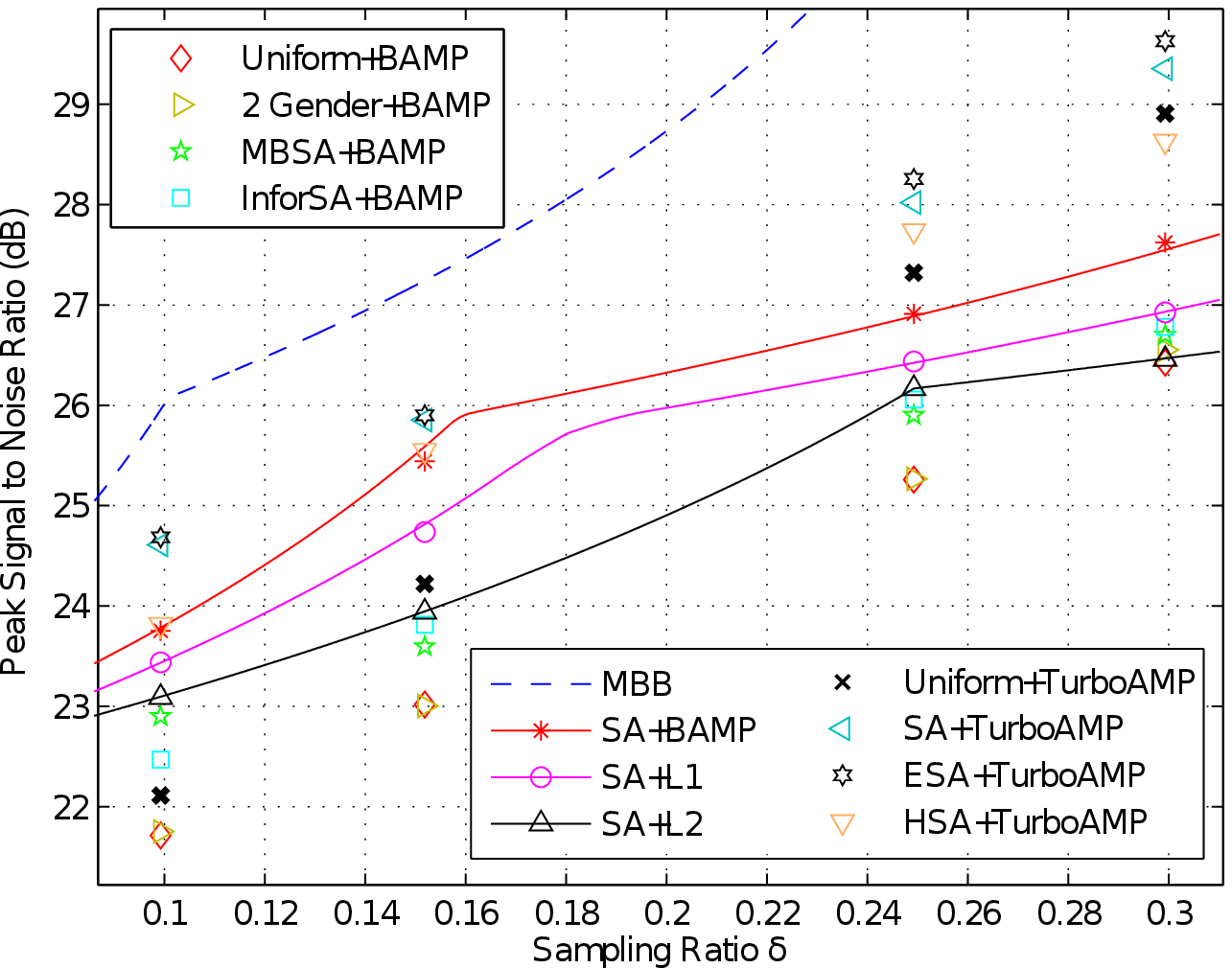}
\caption{PSNR comparison of different encoder-decoder pairs for cameraman Daubechies 2 wavelet with the GMD model. The lines are theoretical predictions with the SD function. While dots represent simulations with the cameraman image.}
\label{fig:SDRdb2GMD}
\end{figure}

The corresponding PSNR results are shown in Fig. \ref{fig:SDRdb2GGD} and Fig. \ref{fig:SDRdb2GMD} for GGD and GMD model,  respectively. The SD function predicts the expected distortion quite accurately for all three choices of the decoder with SA. For both image models, SA achieves the best performance among the five sensing matrices. The advantages of SA over the Uniform matrix and the 2 Gender matrix is significant in spite of the sample ratio. MBSA has a relatively good performance since it has the essence of putting more samples to the coarse bands. Provided with the same image statistics, InforSA tends to allocate more samples to the fine wavelet bands compared with SA. Thus it is not as effective as SA in the low sampling ratio regime. Interestingly the CS scheme, even with an optimized sample allocation, only provides modest reconstruction gains over the classical linear approximation with similarly optimized sample allocation. This actually verifies the discussion in Section \ref{subsec:nwidth}: the rate of decay of error is the same for both the BAMP and $\ltwo$ decoder (though the constants are different). Thus we do not observe overwhelmingly better performance for the BAMP decoder even when SA is performed.

\begin{figure}
 \centering
 \includegraphics[width=8.5cm]{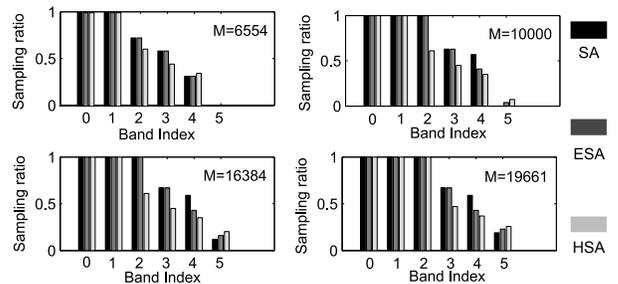}
 \caption{Sample allocation per band for Daubechies 2 wavelet with the GMD model. SA: sample allocation based on the bandwise independent model. HSA: sample allocation based on the empirical SD functions for BAMP decoder with soft information. ESA: empirically optimized sample allocation for turbo decoding.}
 \label{fig:SA}
\end{figure}

Secondly, the quad-tree structure is exploited with the GMD model. Within the turbo decoding regime, simulations are reported for four different sensing matrices: Uniform (amounts to the algorithm proposed in \cite{TurboAMP}), SA, HSA, and the empirically optimized sample allocation, or ESA. As analysed in section \ref{sec:TurboTree}, ESA should be the balance between SA and HSA. For the cameraman image, the ESA is obtained by adaptively reallocating samples from band four to band five based on SA, with the step size of 100 samples, until the PSNR does not increase. The sample allocation per band under four specific sampling ratios are reported in Fig. \ref{fig:SA}. We see that the scaling band and the coarsest wavelet band always have priority over the fine wavelet bands. For this particular image, around 2000 samples are reallocated to the finest scale band to achieve the ESA. For the turbo decoding, the soft information in \eqref{eq:ExInfo} is used. It is fixed if band $j$ is fully sampled during the HMT decoding. For partially sampled bands, activity rates $\lambda_j$ in Table \ref{table:ImPara} are used to initialize the turbo decoding and updated by the HMT decoding for each turbo iteration. Other hyperparameters to initialize the HMT decoding are set in accordance with the recommendation in \cite{TurboAMP}. For various choices of sample allocations, we ran 20 turbo iterations, within which 500 BAMP iterations are performed.

As evident in Fig. \ref{fig:SDRdb2GMD}, adding the HMT decoding ingredient indeed improves the reconstruction quality. it is the joint use of optimized bandwise sampling  and the tree structure that delivers by far the best PSNR performance. Again, sample allocation shows its importances when there is a tight budget of samples: even without the turbo decoding procedure, SA+BAMP is 1.5 dB better at $\delta=0.1, 0.15$ than Uniform+TurboAMP. In the large sampling ratio regime $\delta=0.3$, the effectiveness of the sample allocation is not as obvious and the HMT alone is responsible for the excellent performance: SA+TurboAMP is 0.5 dB better than the Uniform+TurboAMP. It shows that both sample allocation and the HMT play a role in improving the performance of compressive imaging, and which matters more depends on several factors, including the sampling ratio. We also observe that the ESA is only slightly better than the SA. It means that even when we have the luxury of manipulating samples, the benefit is limited because of the exponential energy decay of the multi-resolution model.

The $256\times 256$ cameraman image along with the reconstructed images by different encoder-decoder pairs are visualized in Fig. \ref{fig:cman} at the sampling ratio $\delta=15\%$. It further confirms that given accurate image statistics, our proposed SA is the optimal distribution of samples.

\begin{figure}[t!]
\centering
\includegraphics[width=0.45\textwidth]{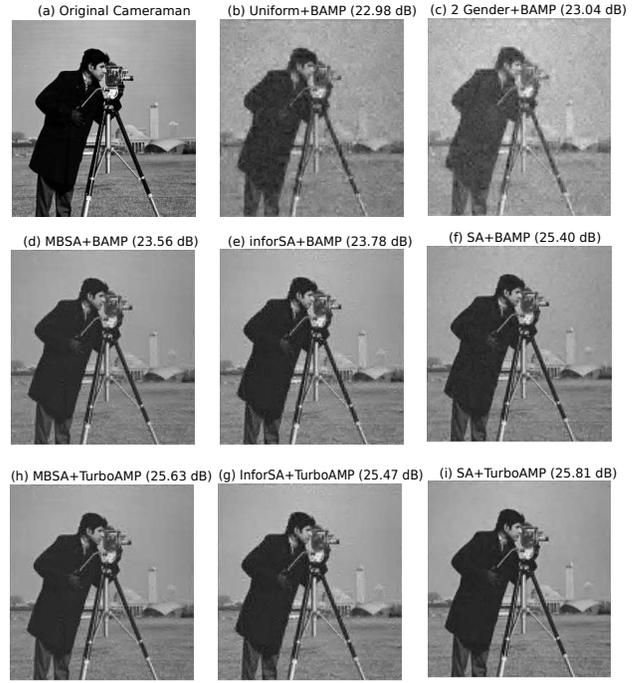}
\caption{Reconstruction using 10000 ($15\%$) samples of the $256\times 256$ cameraman image with different encoder-decoder pairs. The GMD is used to model the Daubechies 2 wavelet coefficients statistics. }
\label{fig:cman}
\end{figure}

\subsection{Sample Allocation with General Image Statistics: the GSA}
In practice, we may not have access to the accurate image statistics. In this section, reconstruction results for a general sample allocation (GSA) which is not tuned to a specific image distribution is presented. The GSA is designed based on the fixed per-band natural image statistics. We estimated the GMD statistics for the six-band Daubechies 2 wavelet decomposition of 200 training images from the \textit{Berkeley Segmentation Dataset} \cite{Berkeley}. Each training image is cropped to the size of $256\times 256$. The pixel intensity value is normalized between 0 and 1. The average per-band GMD parameters are reported in Table \ref{table:AverImPara} and used to generate the general (albeit dictionary and algorithm dependent) sample allocation profile.  

\begin{table}[h!]
\caption{Average Statistics for Daubechies 2 wavelet coefficients of 200 test images from the Berkeley dataset \cite{Berkeley}}\label{table:AverImPara}
\vspace{-0.4cm}
\begin{center}
\resizebox{0.5\textwidth}{!}{
\begin{tabular}{c||cccccc}\hline
subband& $b_1$ & $b_2$ & $b_3$ & $b_4$ & $b_5$\\ \hline \hline
$\lambda$ &0.5108 & 0.4374 & 0.4076 & 0.3616 & 0.3137 \\ \hline
$\sigma_{_L}^2$ & 3.6910 & 0.7506 & 0.1595 & 0.0385 & 0.0081 \\ \hline
$\sigma_{_S}^2$ & 0.4596 & 0.0490 & 0.0075 & 0.0015 & 0.0003\\ \hline
\end{tabular}}
\end{center}
\end{table}

The resulting GSA is then applied to ten test images outside the training set, and again compared with the Uniform matrix, the 2 Gender matrix, MBSA and InforSA. Table \ref{table:AverImPara} is also used to generate InforSA. The BAMP decoder is used as the reconstruction algorithm. The PSNR performance for sampling ratio $\delta=0.1, 0.2, 0.3$ are reported in Table \ref{table:PSNR0_1}, Table \ref{table:PSNR0_2} and Table \ref{table:PSNR0_3}, respectively. 

\begin{figure}[t!]
\centering
\includegraphics[width=0.5\textwidth]{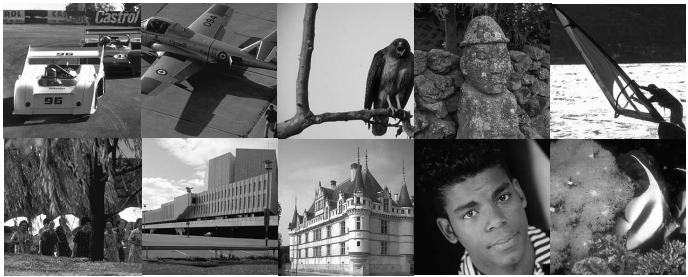}
\caption{Ten test images from the Berkeley dataset \cite{Berkeley}. From left to right, top to bottom are: car, plane, eagle, sculpture, surfer, tourists, building, castle, man and fish.}
\label{fig:cman}
\end{figure}
The reconstruction quality of GSA depends on the accuracy of the image statistics. We see that with reasonable image statistics estimation, GSA outperforms the Uniform matrix and the 2 Gender matrix with roughly 2 dB gain consistently for all cases. The MBSA and InforSA have comparable yet slightly worse performance except three images at sampling ratio $\delta=0.3$. It is due to the actual image deviates from the average image statistics. Not surprisingly, adding the HMT decoding component can only improve the reconstruction quality, if not significantly. 

\section{Conclusion}
\label{sec:5}
The main contribution of this paper is to understand the nature of the sampling for multi-resolution images. For this, the complete sample distortion framework with the definition, lower bounds and the convex property is presented. Given the image statistics, we have derived a tractable sample allocation method for minimizing the reconstruction distortion and shown that it provides an accurate prediction of the achievable SD performance. We have also shown that when the optimized sample allocation is performed, the reconstruction gain of the CS decoder is limited over the linear reconstruction techniques. To get closer to the model based bound, we have deployed the tree structured sparsity within the optimized bandwise sampling framework by the turbo decoding approach. Various encoder-decoder combinations examined with the cameraman image illustrate the merit of bandwise sampling, especially in the regime of very low sampling ratios. For practical sample allocation, a general sampling profile is constructed based on average image statistics and demonstrates competitive performance. 

\begin{table}[h!]
\caption{Image reconstruction results for ten $256\times 256$ test images from the berkeley image database \cite{Berkeley} with $\delta=0.1$. Entries are the peak signal-to-noise ratio (PSNR) in decibels, $\text{PSNR} := 10\log_{10}(N/||\hat{x}-x||^2_2)$. All results use the aveargae image statistics reported in Table \ref{table:AverImPara} and the BAMP decoder.} \label{table:PSNR0_1}
\vspace{-0.4cm}
\begin{center}
\resizebox{0.5\textwidth}{!}{
\begin{tabular}{l|| c|c|c|c|c||c}
\hline
Image & GSA & InforSA & MBSA & Uniform & 2 Gender & SA+TurboAMP \\
\hline
\hline
car & $\mathbf{22.52}$ & 21.67 & 22.28 & 20.61 & 20.65 & $\mathsf{23.12}$ \\ \hline
plane & $\mathbf{25.87}$ & 25.27 & 25.63 & 24.16 & 24.26 & $\mathsf{26.57}$ \\ \hline
eagle & $\mathbf{25.23}$ & 24.53 & 24.88 & 23.39 & 23.44 & $\mathsf{26.30}$ \\ \hline
sculpture & $\mathbf{22.42}$ & 21.72 & 22.36 & 20.75 & 20.81 & $\mathsf{22.68}$ \\ \hline 
surfer & $\mathbf{22.37}$ & 21.58 & 22.11 & 20.42& 20.59 & $\mathsf{23.14}$ \\ \hline
tourists & $\mathbf{22.17}$ & 21.35 & 22.08 & 20.41 & 20.50 & $\mathsf{22.52}$ \\ \hline
building & $\mathbf{22.01}$ & 21.42 & 21.84 & 20.39 & 20.41 & $\mathsf{22.73}$ \\ \hline
castle & $\mathbf{21.40}$ & 20.93 & 21.26 & 19.82 & 19.78 & $\mathsf{21.74}$ \\ \hline
man & $\mathbf{26.86}$ & 26.02 & 26.42 & 24.84 & 24.89 & $\mathsf{28.52}$ \\ \hline
fish & $\mathbf{24.60}$ & 23.52 & 24.43 & 22.57 & 22.63 & $\mathsf{24.85}$ \\ \hline
\end{tabular}}
\end{center}
\end{table}

\begin{table}[h!]
\caption{Reconstruction PSNR for test images with $\delta=0.2$}\label{table:PSNR0_2}
\vspace{-0.4cm}
\begin{center}
\resizebox{0.5\textwidth}{!}{
\begin{tabular}{l|| c|c|c|c|c||c}
\hline
Image & GSA & InforSA & MBSA & Uniform & 2 Gender & SA+TurboAMP \\
\hline
\hline
car & $\mathbf{25.56}$ & 24.11 & 25.29 & 22.92 & 22.98 & $\mathsf{25.92}$ \\ \hline
plane & $\mathbf{28.28}$ & 27.32 & 28.13 & 26.19 & 26.25 & $\mathsf{28.52}$ \\ \hline
eagle & $\mathbf{28.66}$ & 27.84 & 28.59 & 26.31 & 26.44 & $\mathsf{28.95}$ \\ \hline
sculpture & $\mathbf{23.81}$ & 22.89 & 23.54 & 22.05 & 22.61 & $\mathsf{24.58}$ \\ \hline 
surfer & $\mathbf{25.37}$ & 24.00 & 25.13 & 22.81& 22.95 & $\mathsf{25.65}$ \\ \hline
tourists & $\mathbf{24.15}$ & 22.93 & 23.75 & 22.08 & 22.37 & $\mathsf{24.53}$ \\ \hline
building & $\mathbf{24.84}$ & 23.59 & 24.66 & 22.48 & 22.55 & $\mathsf{25.37}$ \\ \hline
castle & $\mathbf{23.65}$ & 22.76 & 23.41 & 21.02 & 21.42 & $\mathsf{23.96}$ \\ \hline
man & $\mathbf{30.32}$ & 29.33 & 30.08 & 28.05 & 28.49 & $\mathsf{30.80}$ \\ \hline
fish & $\mathbf{27.26}$ & 27.57 & 26.76 & 24.62 & 24.83 & $\mathsf{27.76}$ \\ \hline
\end{tabular}}
\end{center}
\end{table}

\begin{table}[h!]
\caption{Reconstruction PSNR for test images with $\delta=0.3$}\label{table:PSNR0_3}
\vspace{-0.4cm}
\begin{center}
\resizebox{0.5\textwidth}{!}{
\begin{tabular}{l|| c|c|c|c|c||c}
\hline
Image & GSA & InforSA & MBSA & Uniform & 2 Gender & SA+TurboAMP \\
\hline
\hline
car & 26.21 & $\mathbf{26.24}$ & 26.15 & 25.00 & 25.22 & $\mathsf{26.97}$ \\ \hline
plane & 28.96 & $\mathbf{29.20}$ & 28.89 & 28.21 & 28.54 & $\mathsf{29.82}$ \\ \hline
eagle & $\mathbf{29.97}$ & 29.22 & 29.17 & 28.61 & 28.94 & $\mathsf{30.25}$ \\ \hline
sculpture & 24.94 & 23.93 & $\mathbf{25.02}$ & 23.00 & 23.11 & $\mathsf{25.72}$ \\ \hline 
surfer & $\mathbf{26.04}$ & 25.96 & 25.85 & 24.91& 25.05 & $\mathsf{26.85}$ \\ \hline
tourists & $\mathbf{25.35}$ & 24.22 & 25.15 & 23.35 & 23.57 & $\mathsf{25.79}$ \\ \hline
building & $\mathbf{25.50}$ & 25.28 & 25.32 & 24.28 & 24.42 & $\mathsf{26.17}$ \\ \hline
castle & $\mathbf{24.32}$ & 24.21 & 24.16 & 23.04 & 23.06 & $\mathsf{24.75}$ \\ \hline
man & $\mathbf{31.56}$ & 30.85 & 30.77 & 30.05 & 30.29 & $\mathsf{33.09}$ \\ \hline
fish & $\mathbf{28.76}$ & 27.97 & 28.26 & 26.31 & 26.53 & $\mathsf{29.31}$ \\ \hline
\end{tabular}}
\end{center}
\end{table}

\appendices
\section{Proof of Theorem \ref{T:EBB}}
\label{app:EBB}
Without loss of generality we will assume that $\boldsymbol{\Phi}$ is an orthogonal projection operator and we denote by $\boldsymbol{\Phi}^{\bot}$ the orthogonal projection onto the null space of $\boldsymbol{\Phi}$. We can then split the signal $\mathbf{x}$ into its observed and unobserved components: $\mathbf{y}=\boldsymbol{\Phi}\mathbf{x}$ and $\mathbf{z}=\boldsymbol{\Phi}^{\bot}\mathbf{x}$. Since we directly observe $\mathbf{y}$ we need only consider the component of the decoder that estimates $\mathbf{z}$, $\boldsymbol{\Delta}^{(\mathbf{z})}:\R^m\to\R^{n-m}$. We can then estimate $\mathbf{x}$ as:
\begin{equation}
\hat{\mathbf{x}}=\boldsymbol{\Delta}(\mathbf{y})=\boldsymbol{\Phi}^T \mathbf{y}+[\boldsymbol{\Phi}^{\bot}]^T\boldsymbol{\Delta}^{(\mathbf{z})}(\mathbf{y})
\end{equation}
We can further write the squared error distortion in terms of $\boldsymbol{\Delta}^{(\mathbf{z})}(\mathbf{y})$ as
\begin{equation}
D=\frac{1}{n}\int\ p(\mathbf{y})\int\ p(\mathbf{z}|\mathbf{y})||\mathbf{z}-\boldsymbol{\Delta}^{(\mathbf{z})}(\mathbf{y})||^2_2\ \mathrm{d}\mathbf{z}\mathrm{d}\mathbf{y}
\end{equation}

Now consider the following decomposition of the differential entropy $h(\mathbf{x})$ of the vector $\mathbf{x}$:
\begin{eqnarray}
\label{eq:EBB}
h(\mathbf{x})&=& h(\mathbf{y})+h(\mathbf{z}|\mathbf{y}) \nonumber\\
    &=& h(\mathbf{y})+h(\mathbf{z}-\boldsymbol{\Delta}^{(\mathbf{z})}(\mathbf{y})|\mathbf{y}) \nonumber\\
    &\leq & h(\mathbf{y})+h(\mathbf{z}-\boldsymbol{\Delta}^{(\mathbf{z})}(\mathbf{y}))\nonumber \\
    &\leq & \frac{m}{2} \log_2 2\pi e+\frac{n-m}{2}\log_2 2\pi enD/(n-m) \nonumber\\
\end{eqnarray}
where we have used the following observations
\begin{itemize}
\item (line 2) The decoder is a deterministic function of $\mathbf{y}$ and therefore the differential entropy of $h(\mathbf{z}-\boldsymbol{\Delta}^{(\mathbf{z})}(\mathbf{y})|\mathbf{y})=h(\mathbf{z}|\mathbf{y})$.
\item (line 3) The conditional entropy is bounded by the marginal entropy: $h(\mathbf{x}|\mathbf{y})\leq h(\mathbf{x})$.
\item (line 4) The entropy of a random variable with a fixed covariance is bounded by the entropy of a Gaussian with the same covariance. Similarly the entropy of a random vector $\mathbf{v}\in\R^{n-m}$ under the constraint that $\Expect\lbrace \mathbf{v}^T\mathbf{v}\rbrace=nD$ is bounded by the entropy of a Gaussian random vector with covariance $\frac{nD}{(n-m)I}$.
\end{itemize}

The principle here is that the optimal projection should maximize the entropy of the observed component $h(\mathbf{y})$ while the decoder, $\boldsymbol{\Delta}(\mathbf{y})$, should minimize the distortion possible. This is similar to the concept of information sensing proposed in \cite{infosensing}.

Substituting $\delta=m/n$ into \eqref{eq:EBB} gives:
\begin{equation}
h(x)\leq \frac{1-\delta}{2}\log_2 2\pi e\frac{D}{1-\delta}+\frac{\delta}{2}\log_2 2\pi e
\end{equation}
where we have used the i.i.d assumption to write $h(\mathbf{x})=nh(x)$. This can then be rearranged to give the EBB.

\section{Derivation of the Hierarchical Bayesian Model for the GGD}
\label{app:MBB}
Here, we derive the hierarchical Bayesian model to describe the GGD, which is then used to bound the MSE performance described in the main text in Sec. \ref{sec:LB}. We introduce two latent variables $c_1$ and $c_2$ to simplify the expression of GGD:
\begin{equation}
c_1 =\frac{\alpha}{2\sqrt{\beta}\sigma\Gamma(\frac{1}{\alpha})} \quad  c_2 =(\sqrt{\beta}\sigma)^{\alpha}
\end{equation}

Then the pdf of GGD can be written as 
\begin{equation}
p_{_\textrm{GGD}}(x)=c_1 \textrm{exp}(-\frac{|x|^\alpha}{c_2})
\end{equation}
Let $p(x|\tau)=\N(x;0,\tau)$. To establish the hierarchical model, we need to find the prior $p(\tau)$ which satisfies:
\begin{equation}
\int_0^\infty \! \N(x;0,\tau)p(\tau) \, \mathrm{d}\tau=c_1 \textrm{exp}(-\frac{|x|^\alpha}{c_2})
\end{equation}
Using the substitution $g(\tau)=\frac{1}{\sqrt{2\pi\tau}}p(\tau)$, $m=\frac{x^2}{2}$ and $t=\frac{\sqrt{2}^\alpha}{c_2}$, the question becomes solving $g(\tau)$ subject to
\begin{equation}
\int_0^\infty \! \textrm{exp}(-\frac{m}{\tau})g(\tau) \, \mathrm{d}\tau=c_1 \textrm{exp}(-tm^{\frac{\tau}{2}})
\end{equation}
let $z=\frac{1}{\tau}$ and $G(z)=g(\tau)|_{\tau=\frac{1}{z}}$, we further transform the problem to find $G(z)$ subject to
\begin{equation}
\int_0^\infty \! \textrm{exp}(-zm)\frac{G(z)}{z^2} \, \mathrm{d}z=c_1 \textrm{exp}(-tm^{\frac{\tau}{2}})
\end{equation}
Applying the integral formula \cite{Polyanin}: if $\int_0^\infty \! e^{-zt}y(t) \, \mathrm{d}t=f(z)$, then $y(t)=\mathcal{L}^{-1}(f(z))$, we obtain
\begin{equation}
\label{ILGGD}
\frac{G(z)}{z^2}=c_1 \mathcal{L}^{-1}\textrm{exp}(-\frac{x^\alpha}{c_2})]
\end{equation}
where $\mathcal{L}^{-1}(\cdot)$ is the inverse Laplace transform. The inversion of Laplace transform in (\ref{ILGGD}) can be solved numerically \cite{IL}. From here we obtain the MBB for the GGD data in Fig. \ref{fig:sdggd}.




%



\bibliographystyle{IEEEtran}
\bibliography{SAtreeBib}

%








\end{document}